\documentclass[10pt,twocolumn,letterpaper,pagenumbers]{article}

\usepackage{iccv}              
\usepackage{bm}
\usepackage{multirow}
\usepackage[T1]{fontenc}            
\usepackage{newtxtext,newtxmath}    
\usepackage[final]{microtype}
\usepackage{fix-cm}                 
\definecolor{iccvblue}{rgb}{0.21,0.49,0.74}
\usepackage[pagebackref,breaklinks,colorlinks,allcolors=iccvblue]{hyperref}

\def\paperID{8} 
\def\confName{ICCV}
\def\confYear{2025}

\title{SlimComm: Doppler-Guided Sparse Queries for Bandwidth-Efficient Cooperative 3-D Perception}

\author{%
  Melih Yazgan\textsuperscript{1}\textsuperscript{†}\quad
  Qiyuan Wu\textsuperscript{1,2}\textsuperscript{†}\quad
  Iramm Hamdard\textsuperscript{1,2}\quad
  Shiqi Li\textsuperscript{1,2}\quad
  J.~Marius Zoellner\textsuperscript{1,2}\\
  \textsuperscript{1}FZI Research Center for Information Technology, 
  \textsuperscript{2}Karlsruhe Institute of Technology\\
  \footnotesize\texttt{last.name@fzi.de}\\\footnotesize\textsuperscript{†}These authors contributed equally.
}

\begin{document}
\maketitle

\begin{abstract}
Collaborative perception allows connected autonomous vehicles (CAVs) to overcome occlusion and limited sensor range by sharing intermediate features. Yet transmitting dense Bird’s-Eye-View (BEV) feature maps can overwhelm the bandwidth available for inter-vehicle communication. We present SlimComm, a communication-efficient framework that integrates 4D radar Doppler with a query-driven sparse scheme. SlimComm builds a motion-centric dynamic map to distinguish moving from static objects and generates two query types: (i) reference queries on dynamic and high-confidence regions, and (ii) exploratory queries probing occluded areas via a two-stage offset. Only query-specific BEV features are exchanged and fused through multi-scale gated deformable attention, reducing payload while preserving accuracy. For evaluation, we release OPV2V-R and Adver-City-R, CARLA-based datasets with per-point Doppler radar. SlimComm achieves up to 90\% lower bandwidth than full-map sharing while matching or surpassing prior baselines across varied traffic densities and occlusions. Dataset and code will be available at: \url{https://url.fzi.de/SlimComm}.
\end{abstract}
    
\section{Introduction}
\label{sec:intro}
Autonomous driving and other unmanned systems have made rapid strides, spurred by the demand for reliable 3-D object detection~\cite{liu_bevfusion_2023}. Yet dependable perception in complex outdoor environments remains difficult because of occlusions and sensor-specific limitations~\cite{9578621}. LiDAR supplies centimetre-level geometry but degrades at long range~\cite{9760734}; conversely, 4-D radar offers resilient range–Doppler measurements with coarser angular resolution~\cite{9760734,9564754}. Adding elevation to conventional radar returns enables preprocessing similar to LiDAR point clouds~\cite{wang_interfusion_2022,lin2024rcbevdet}.

Collaborative perception through Vehicle-to-Everything (V2X) communication can overcome these weaknesses by allowing vehicles to see beyond their own field of view. However, broadcasting dense Bird’s-Eye-View (BEV) feature maps quickly overwhelms the limited bandwidth available for inter-vehicle communication~\cite{yazgan_cope_survey}. The key question is therefore which information to share and from whom to request it.

We answer this question with SlimComm, a proactive cooperative-perception framework that transmits only a sparse set of high-value queries instead of full feature maps. The query strategy follows two steps:
\begin{itemize}
    \item[\textit{(i)}] \textit{Dynamic map.} Ego-motion-compensated radar Doppler forms a motion-centric map; heuristic queries placed on this map focus on moving objects.
    \item[\textit{(ii)}] \textit{Exploratory queries.} A confidence prior highlights likely foreground cells; additional queries dropped in the resulting occlusion shadows prompt collaborators to recover partially or fully hidden objects.
\end{itemize}
Each ego vehicle broadcasts its queries. Chosen neighbours warp their local BEV features into the ego frame, extract a halo-enriched context window around every query, and return those features. The ego agent then fuses ego and neighbour responses using a gated multi-scale deformable-attention block.
\paragraph{Contributions.}
\begin{itemize}
    \item Doppler-guided semantic query generation: a two-step radar-guided query strategy that selectively targets moving and occlusion-prone regions, yielding a superior accuracy–bandwidth trade-off.
    \item Communication-efficient collaborator selection: a lightweight neighbour-selection scheme that transmits only a 3$\times$3 halo of BEV features per query, reducing transmitted bytes to $\sim$10\% of full-map sharing.
    \item OPV2V-R and Adver-City-R benchmarks: CARLA-based extensions of two V2X suites with synchronised LiDAR–4-D radar, released alongside a full training code to catalyse reproducible multi-agent Doppler research.\footnote{A real-world V2X-Radar dataset has been reported in~\cite{yang2024v2x}, but the data were not released at the time of writing.}
\end{itemize}

\section{Related Work}\label{sec:related}
\subsection{Sensing Modalities}
Most cooperative perception systems rely on LiDAR alone or LiDAR–camera fusion~\cite{hu_where2comm_2022,xu_opv2v_2022,qiao2023adaptive,liu_when2com_2020,Xu_2025_CVPR,wang_v2vnet_2020,yang2023spatio,10946103,cosense3d,8885377}.
In contrast, LiDAR–radar fusion has recently advanced single-vehicle perception: LiRaFusion introduces a learnable gating scheme~\cite{song2024lirafusion}; InterFusion employs pillar-wise attention~\cite{wang_interfusion_2022}; RLNet compensates for radar noise~\cite{xu2025rlnet}; and Bi-LRFusion couples the modalities bidirectionally~\cite{Wang2023BiLRFusion}. However, none of these approaches addresses inter-vehicle occlusions or bandwidth limitations in multi-agent scenarios. Incorporating radar Doppler into a cooperative framework thus remains largely unexplored.

\subsection{Communication-Efficient Cooperative Perception}
Early frameworks such as V2VNet~\cite{wang_v2vnet_2020}, Attentive Fusion~\cite{xu_opv2v_2022}, and AdaFusion~\cite{qiao2023adaptive} exchange dense BEV feature maps among CAVs. While this maximises perception accuracy, it quickly overloads the V2X channel~\cite{yazgan_cope_survey}. Compressing these maps reduces channel usage but at the cost of precision.

Subsequent methods therefore focus on sparse communication. Where2Comm~\cite{hu_where2comm_2022} transmits only high-confidence BEV regions; EffiComm~\cite{yazgan2025efficomm} further prunes them via its Adaptive Grid Reduction (AGR) module; When2Com~\cite{liu_when2com_2020} triggers exchange when ego-view uncertainty increases; and CoSDH~\cite{Xu_2025_CVPR} selects messages based on a supply–demand model. SCOPE~\cite{yang2023spatio}, StreamLTS~\cite{cosense3d}, and DelAwareCol~\cite{10946103} incorporate spatio-temporal filtering to further reduce bandwidth.

Despite these advances, most methods still rely on statistical confidence or entropy measures. They rarely encode semantic knowledge of motion or occlusion, resulting in the transmission of large amounts of static background rather than dynamic objects of interest.

\subsection{Cooperative-Perception Datasets}
A comprehensive review of V2X datasets is provided in~\cite{yazgan_dataset_survey}.
OPV2V~\cite{xu_opv2v_2022} and AdverCity~\cite{karvat_adver-city_2025} provide simulated LiDAR–camera data within CARLA, while DAIR-V2X~\cite{yu_dair-v2x_2022} contains real-world LiDAR–camera recordings. More recently, V2X-R~\cite{Huang_2025_CVPR} introduces cooperative radar data, but it omits per-point Doppler information and relies solely on a front-facing sensor, limiting both motion analysis and 360° situational awareness.

\paragraph{Summary.}
Existing methods either broadcast full feature maps to maximise accuracy or reduce bandwidth using sparsity heuristics. They overlook occlusion recovery, which is critical for both perception completeness and efficient collaboration, and they under-utilise radar in multi-agent contexts, partly due to the scarcity of suitable public datasets. In contrast, SlimComm integrates Doppler-aware, occlusion-guided queries with LiDAR–radar cooperation to improve the accuracy–bandwidth trade-off.

\begin{figure}[t!]
  \centering
  \includegraphics[width=1\linewidth]{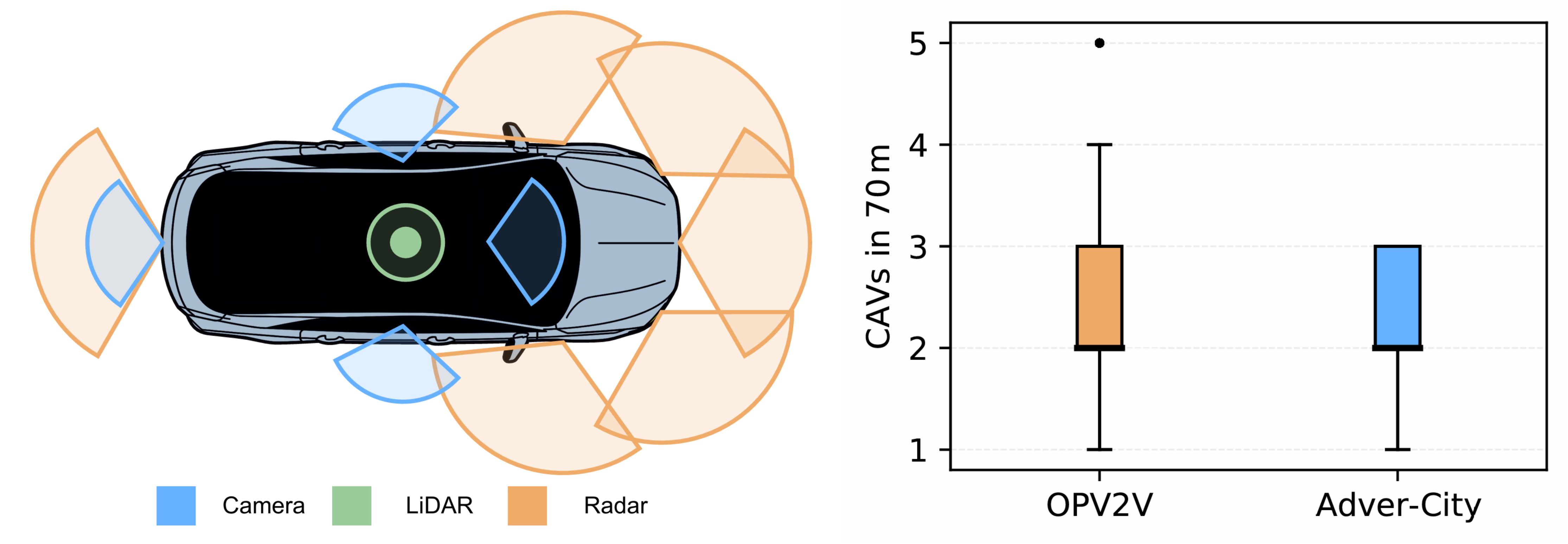}
  \caption{Left: Sensor-suite configuration for each CAV. The six-radar setup is designed to provide full $360^{\circ}$ Doppler velocity coverage, with three radars covering the front, one the rear, and two monitoring adjacent lanes. Right: Number of neighbouring CAVs within \(70\) m per frame.}
  \label{fig:RadarSetup}
\end{figure}
\section{Dataset}
To address the gap in publicly available V2X benchmarks and to properly evaluate multi-agent models that fuse 4-D radar, we introduce radar-augmented versions of the OPV2V~\cite{xu_opv2v_2022} and Adver-City~\cite{karvat_adver-city_2025} datasets. Our goal is to benchmark models that leverage point clouds enriched with Doppler velocity under a variety of driving conditions.  
\begin{table}[b!]
\resizebox{\columnwidth}{!}{%
\begin{tabular}{c|c}
\textbf{Sensor} & \textbf{Specification} \\ \hline
4× Camera & RGB, 800 × 600, 110\textdegree{} FOV \\ \hline
1× LiDAR & 64 ch., 1.3 M pts/s, 120 m, –25\textdegree{}–2\textdegree{} vert.\ FOV, 0.02 m noise, 20 Hz \\ \hline
6× Radar & 0.06 M pts/s, 150 m, 120\textdegree{} horiz.\ FOV, 30\textdegree{} vert.\ FOV \\ \hline
GPS/IMU & GNSS alt.\ noise 0.001 m; IMU heading 0.1\textdegree{}, speed 0.2 m/s
\end{tabular}}
\caption{Sensor suite configuration for OPV2V-R and Adver-City-R.}
\label{tab:SensorSetup}
\end{table}
 \begin{figure*}[h!]
    \centering
    \includegraphics[width=0.90\linewidth]{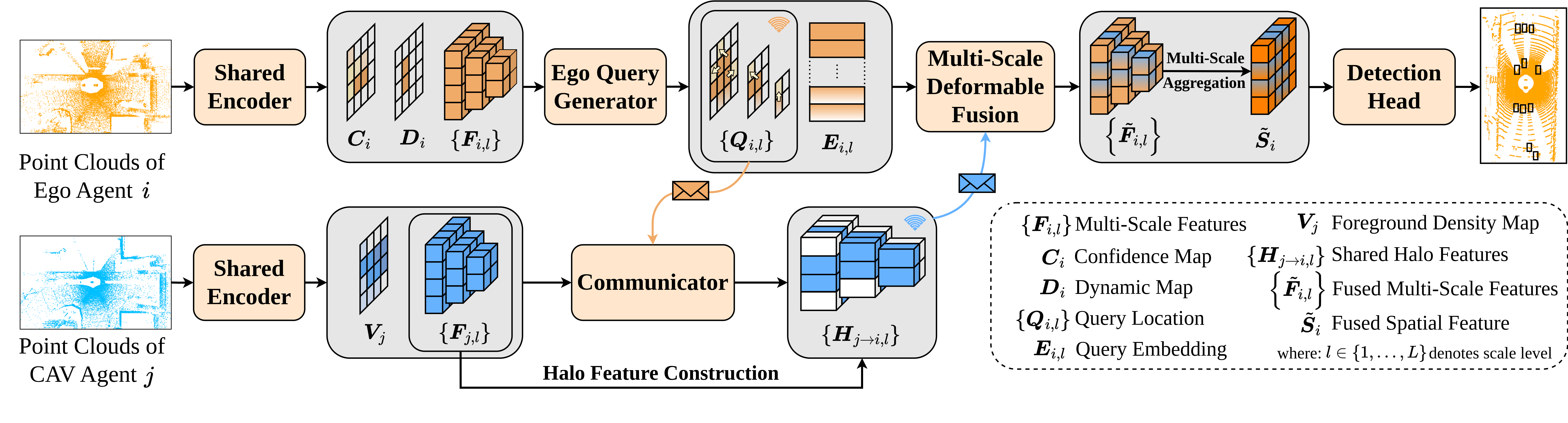}
    \caption{End-to-end cooperative perception. Each agent voxelises its LiDAR–radar points and runs a \textbf{shared encoder} that outputs multi-scale features and three semantic priors (dynamic, confidence, density). The \textbf{Ego Query Generator} uses those priors to emit sparse queries, neighbours respond with \textbf{halo-enriched} features, and a \textbf{gated deformable fusion} module merges everything into a BEV tensor for detection.}
    \label{fig:overall}
\end{figure*}
\subsection{Dataset Creation}\label{sec:dataset}
OPV2V covers generic urban traffic with diverse intersection types, occlusion patterns and flow densities~\cite{xu_opv2v_2022}. Adver-City instead concentrates on crash-relevant situations derived from real accident statistics, including complex junctions and rural roads with restricted sight lines~\cite{karvat_adver-city_2025}. By equipping both suites with an identical sensor package we enable consistent evaluation of radar-enhanced perception across complementary scenarios.

\textbf{OPV2V-R} and \textbf{Adver-City-R} are generated with the same CARLA~\cite{Dosovitskiy17} and OpenCDA~\cite{xu2021opencda} pipeline, recorded at 10Hz, and annotated with fully compatible 3-D bounding boxes. For Adver-City-R, we select the \emph{ClearDay} weather profile, remove roadside units, and harmonise vehicle classes with OPV2V-R (e.g., \ excluding micro-cars). Pedestrian labels are retained but can be ignored during training to match OPV2V-R. \cref{fig:RadarSetup} shows that both datasets have a median of \(\approx\!2\) neighbouring CAVs per frame. OPV2V-R exhibits a broader tail (up to five neighbours), whereas Adver-City-R is more narrowly distributed.

Each CAV carries RGB cameras, a LiDAR, GNSS/IMU, and six simulated radars that output XYZ coordinates and Doppler velocity (see \cref{tab:SensorSetup} for specifications and \cref{fig:RadarSetup} for mounting positions). Cameras provide 360\textdegree{} coverage; the LiDAR is roof-mounted; three radars cover the front bumper, one the rear, and two under the side mirrors look backward to monitor adjacent lanes. Additional dataset statistics and qualitative comparisons are provided in the supplementary material.


\section{Method}
\label{sec:Method}
\subsection{Overall Architecture} \label{sec:overall}
\cref{fig:overall} outlines the end-to-end cooperative-perception pipeline.  
Each agent first feeds voxelised LiDAR and radar data into a shared Encoder, producing multi-scale feature tensors $\{\bm F_{i,l}\!\in\!\mathbb R^{C_l\times H_l\times W_l}\}$.

Alongside these features, the encoder outputs three semantic priors (see \cref{sec:encoder}):  
a Dynamic Map $\bm D_i$, a Confidence Map $\bm C_i$, and a Foreground Density Map $\bm V_i$.  
These maps guide the Ego Query Generator(see~\cref{sec:query_generator}), which converts the down-sampled $\bm D_{i,l}$ and 
 
Each neighbour consults its own density map $\bm V_j$ to decide whether to participate, thereby filtering out uninformative links and reducing bandwidth.
Agents that opt in return halo-enriched features $\{\bm H_{j\!\to\!i,l}\}$ (see \cref{sec:communicator}).  
These responses are merged by the Gated Multi-Scale Deformable Fusion module into per-scale fused features $\{\tilde{\bm F}_{i,l}\}$, which are then aggregated across scales to form a unified BEV tensor $\tilde{\bm S}_i$. A lightweight detection head operating on $\tilde{\bm S}_i$ produces the final 3-D predictions.
Overall, the framework selects only informative collaborators and integrates complementary observations into a compact BEV representation, achieving accurate yet bandwidth-efficient cooperative perception.
\subsection{Encoder}\label{sec:encoder}
\begin{figure}[htbp]
    \centering
    \includegraphics[width=0.8\linewidth]{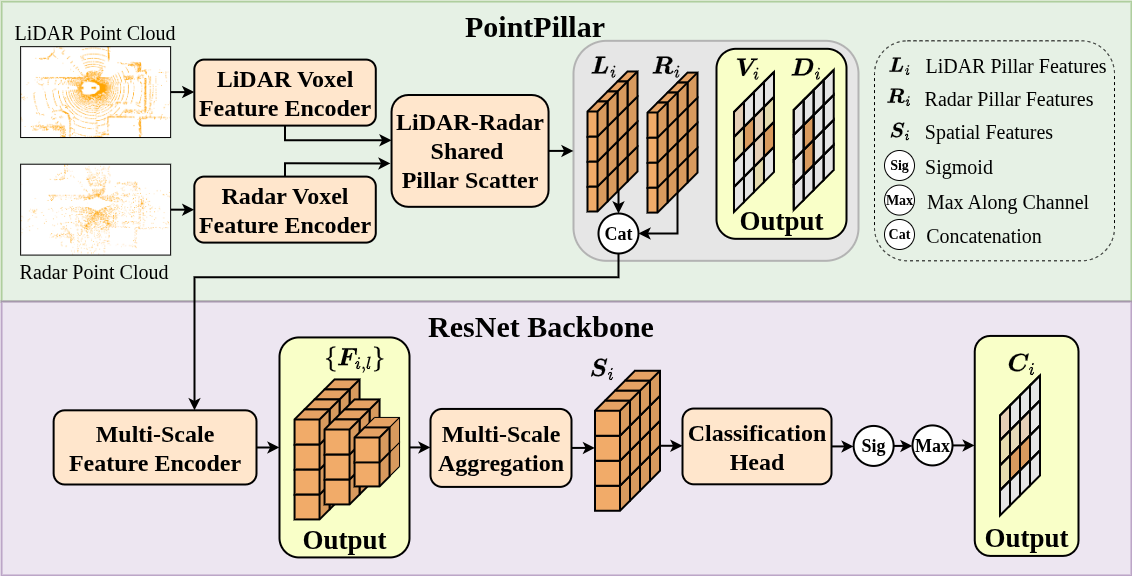}
    \caption{Encoder overview. LiDAR–radar pillars are concatenated in BEV space and processed by a ResNet backbone, which outputs a feature pyramid and three semantic priors.}
    \label{fig:encoder}
\end{figure}

\paragraph{Backbone.}
As shown in  \cref{fig:encoder}, following PointPillars~\cite{lang_pointpillars_2019}, LiDAR and radar point clouds are pillarised into BEV tensors
$\bm L_i$ and $\bm R_i$, concatenated, and fed to a ResNet BEV backbone~\cite{He2015}.
The backbone yields multi-scale features $\{\bm F_{i,l}\}$ and a high-resolution map $\bm S_i$.
A classification head along with sigmoid and channel-wise max operating on $\bm S_i$ produces the confidence map
$\bm C_i\in[0,1]^{H\times W}$.
The next two subsections detail the auxiliary priors used later for query generation and collaborator selection.
$\bm C_{i,l}$ into sparse query locations $\{\bm Q_{i,l}\}$ for deformable attention. Before feature exchange, the ego agent transmits its query locations to neighbouring agents through the Communicator. 
\subsubsection{Dynamic Map}\label{sec:dynamic-map}
Radar Doppler information is converted into a binary motion mask after ego-motion compensation.  
Let $\mathbf v^{\text{veh}}_i$ be the ego velocity of vehicle $i$ in its own frame.
For the $k$-th radar on that vehicle, the velocity in radar coordinates is
\begin{equation}
\mathbf v^{\text{veh}}_{i,k} = \mathbf R_{i,k}\,\mathbf v^{\text{veh}}_i ,
\end{equation}
with $\mathbf R_{i,k}$ the extrinsic rotation matrix.

\begin{figure}[t!]
    \centering
    \includegraphics[width=0.8\linewidth]{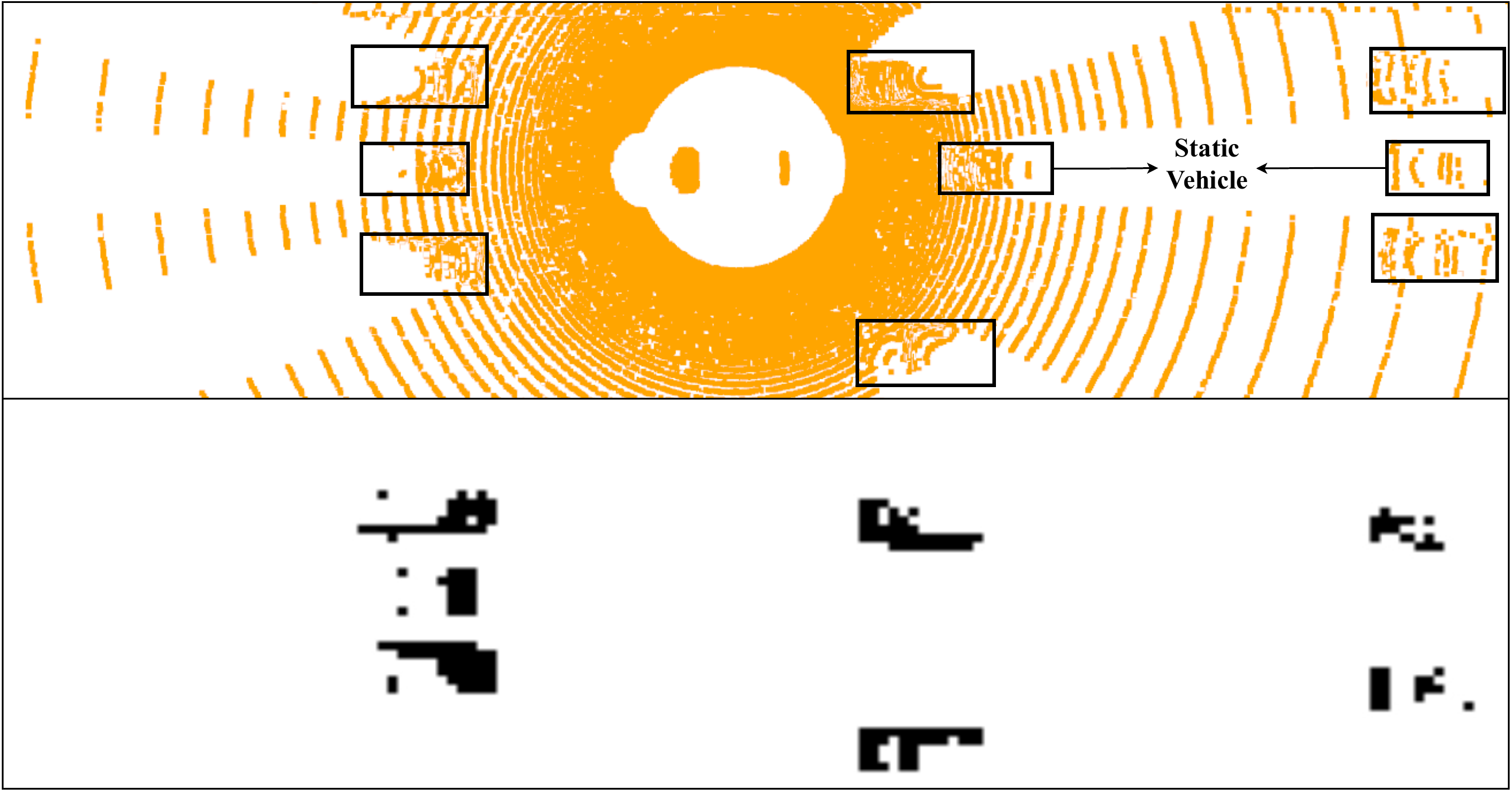}
    \caption{Top: Raw point cloud with GT bounding boxes. Bottom: Dynamic Map. All dynamic vehicles are captured as dynamic grids (black), while static vehicles are excluded from dynamic regions.}
    \label{fig:Dynamic_map}
\end{figure}

The measured Doppler velocity for point $n$ is
\begin{equation}
v^{\text{Doppler}}_{i,k,n}
  =\bigl(\mathbf v^{\text{abs}}_{i,k,n}-\mathbf v^{\text{veh}}_{i,k}\bigr)\!\cdot\!\mathbf u_{i,k,n},
\end{equation}
where $\mathbf u_{i,k,n}=\mathbf p_{i,k,n}/\|\mathbf p_{i,k,n}\|$ is the line-of-sight unit vector.  
Re-arranging gives the compensated radial velocity
\begin{equation}
v^{\text{radial}}_{i,k,n}
  = v^{\text{Doppler}}_{i,k,n}
    + \mathbf v^{\text{veh}}_{i,k}\!\cdot\!\mathbf u_{i,k,n}.
\end{equation}
A BEV cell is marked dynamic if any radar point inside it satisfies
$\lvert v^{\text{radial}}_{i,k,n}\rvert > v_{\text{th}}$,
with $v_{\text{th}}=1.0$ m/s.
This yields the binary Dynamic Map
$\bm D_i\in\{0,1\}^{H\times W}$. ~\cref{fig:Dynamic_map} demonstrates that the dynamic map reliably distinguishes moving and static objects, achieving precise dynamic–static separation.

\subsubsection{Foreground Density Map}\label{sec:fg-density-map}

The Foreground Density Map highlights informative regions by combining height-based foreground masking with point density.

\textbf{Foreground masking.}
Using thresholds
$T_{\text{lower}}\!=\!-1.2$ m,
$T_{\text{upper}}\!=\!0$ m,
and
$T_{\text{max}}\!=\!1.0$ m,
a cell is background if it contains
(i) any point above $T_{\text{max}}$ (tall static structures), or
(ii) all points outside $[T_{\text{lower}},T_{\text{upper}}]$ (ground or noise).
Foreground masks from LiDAR and radar are combined via
\begin{equation}
\bm{FG}_i = \bm{FG}^{\text L}_i \lor \bm{FG}^{\text R}_i .
\end{equation}

\textbf{Density scaling.}
Point counts $\bm N^p_i$ are normalised by the pillar capacity $N_{\text{max}}$:
\begin{equation}
\bm{DS}_i = \bm N^p_i / N_{\text{max}} .
\end{equation}

\textbf{Final map.}
The Foreground Density Map is the element-wise product
\begin{equation}
\bm V_i = \bm{FG}_i \odot \bm{DS}_i .
\end{equation}
It suppresses empty or background cells while retaining dense foreground evidence, and is later used for collaborator selection.

\subsection{Ego Query Generator}
\label{sec:query_generator}
\begin{figure}[t!]
    \centering
    \includegraphics[width=0.80\linewidth]{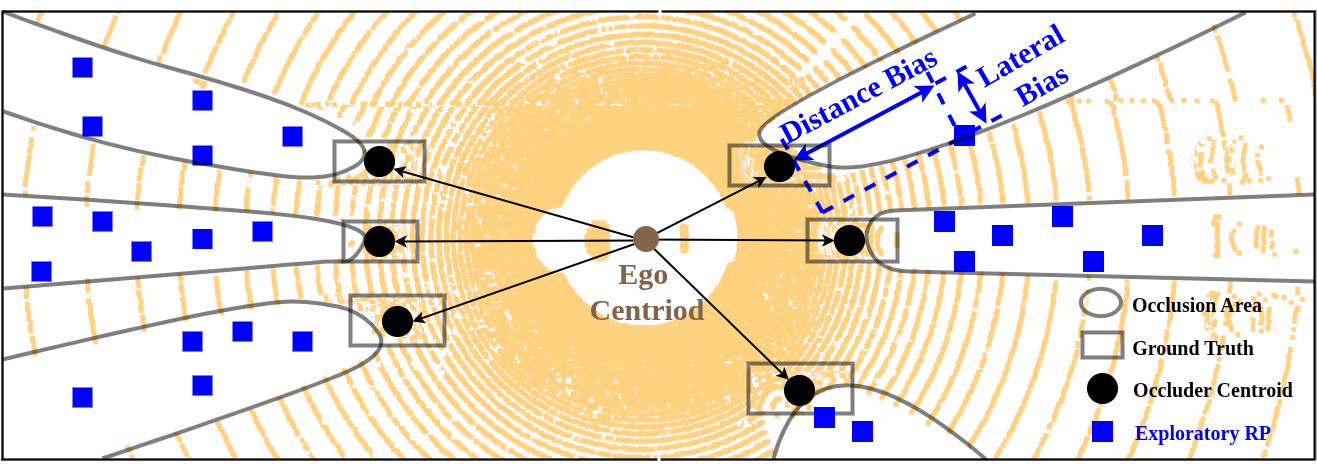}
    \caption{Exploratory reference points are placed behind occluder centroids at a stochastic distance and lateral bias.}
    \label{fig:e_rp}
\end{figure}

To minimise communication overhead, the query generator produces a sparse set of reference points focused on the most informative regions of the scene.
It runs per scale in two stages, yielding Heuristic Reference Points (HRP) for refining visible objects and Exploratory Reference Points (ERP) for probing occluded areas.
\paragraph{Heuristic Branch.}
This branch focuses on regions that already exhibit strong object evidence.
For each scale \(l\), the dynamic and confidence maps are down-sampled,
\(\bm D_{i}\!\rightarrow\!\bm D_{i,l}\) and \(\bm C_{i}\!\rightarrow\!\bm C_{i,l}\),
and HRP locations are drawn from two candidate pools:
\begin{enumerate}
    \item every grid cell flagged as dynamic in \(\bm D_{i,l}\);
    \item the highest-scoring cells in \(\bm C_{i,l}\) that are not in Pool 1, selected until the per-scale budget \(N_l^{r}\) is met.
\end{enumerate}
The resulting set,
\(\bm R_{i,l}^{h}\!\in\!\mathbb R^{N_l^{r}\times2}\),
stores BEV coordinates \((u,v)\).
Embeddings
\(\bm E_{i,l}^{h}\!\in\!\mathbb R^{N_l^{r}\times C_l}\)
are obtained by bilinearly sampling the ego BEV feature map, yielding a rich, context-aware starting point for object refinement.
\paragraph{Exploratory Branch.}
Occlusions often hide critical objects; this branch explicitly seeks them.
\begin{enumerate}
    \item \textbf{Occluder identification.}
         Significant peaks in the confidence map are extracted as occluder centroids,
         \begin{equation}
           \bm C_{i,l}^{o} = \operatorname{MaxPool}_{3\times3}(\bm C_{i,l}) \;\wedge\; \bm C_{i,l},
         \end{equation}
         where \(\wedge\) denotes element-wise logical AND. Border pixels are first masked out
         to suppress artificial edges. A per-scene percentile threshold is then applied.
    \item \textbf{Shadow sampling.}
         For each centroid, a shadow ERP is placed at a stochastic, biased offset, forming
         \(\bm R_{i,l}^{e}\!\in\!\mathbb R^{N_l^{r}\times2}\) (see \cref{fig:e_rp}).
    \item \textbf{Contextual embedding.}
         Each ERP embedding concatenates three signals:
         (i) the occluder’s BEV feature,
         (ii) the 2-D shadow offset, and
         (iii) a scale-specific learnable exploration token \(\bm t_l\).
         The token \(\bm t_l\in\mathbb R^{C_l}\) is a single parameter vector
         shared by all ERPs at level \(l\),
         initialised with Xavier normal noise and updated end-to-end during training.
         The concatenated vector is passed to a two-layer MLP,
         yielding \(\bm E_{i,l}^{e}\in\mathbb R^{N_l^{r}\times C_l}\).
\end{enumerate}
\paragraph{Query aggregation.}
For each scale \(l\), we concatenate the HRP and ERP into an anchor set
\begin{equation}
A_{i,l}=R^{\mathrm{H}}_{i,l}\cup R^{\mathrm{E}}_{i,l},\qquad
|A_{i,l}|=2N^{r}_{l}.
\end{equation}

\textbf{Two-stage offset strategy.}
A lightweight MLP predicts one coarse 2-D
offset for every anchor, forming the nudged centres
\(\tilde A_{i,l}=A_{i,l}+O_{i,l}\).

\textbf{Query-offset regularisation.}
To prevent the HRP and ERP branches from collapsing into identical
behaviour, we introduce an auxiliary loss that encourages exploratory
offsets to be larger than heuristic ones by a scale-dependent
margin \(\delta_l\).  Let
\(O^{\mathrm{H}}_{i,l}\) and \(O^{\mathrm{E}}_{i,l}\) denote the two
offset subsets; the loss reads
\begin{equation}
\mathcal{L}_{\text{offset}}
=\sum_{l}
\Bigl[
\delta_l-
\bigl(\mathbb{E}\lVert O^{\mathrm{E}}_{i,l}\rVert_2-
       \mathbb{E}\lVert O^{\mathrm{H}}_{i,l}\rVert_2\bigr)
\Bigr]_{+},
\end{equation}
with \([\cdot]_{+}\) the ReLU and
\(\delta_l\)
the average occluder-to-shadow distance at that scale.
This term is added to the main detection objective with a dynamic
weight proportional to the primary PointPillar detection loss value.

\textbf{Fine sampling.}
Each nudged centre \(\tilde a\in\tilde A_{i,l}\) is passed to
Deformable Attention, which predicts a learnable
\(3\times 3\) halo (\(n_{\text{points}}=9\)) of fine offsets,
yielding the final sampling locations used to gather collaborators
features. The final set of sampling locations used to gather collaborator features is then given by:
\begin{equation}
\bm{Q}_{i,l} = { \tilde{a} + \Delta^{\text{fine}}_{h,p} \mid \tilde{a} \in \tilde{A}_{i,l}, \forall h, p }.
\end{equation}
The two-stage design lets the network first learn a global
(coarse) correction and then a local, detail-oriented sampling pattern
(fine) without incurring additional bandwidth. 

We further examine the spatial distributions of HRP and ERP to confirm that they capture complementary patterns (visible objects vs.\ occluded regions). 
Qualitative results are provided in the supplementary material.
\subsection{Communicator} \label{sec:communicator}
To enable efficient and adaptive multi-agent perception, the module first selects collaborators and then transmits only sparse, halo-enriched features that are spatially aligned with the ego’s query locations.
\paragraph{Collaborator Selection.}\ 
Prior to feature exchange, the ego broadcasts its query locations \(\bm Q_{i,l}\) and global pose to neighbouring agents.  
Because these queries are defined in the ego coordinate frame, each candidate agent \(j\) warps its Foreground Density Map \(\bm V_j\) into that frame, yielding \(\bm V_{j\!\to\!i}\).
An agent becomes a collaborator whenever at least one query lands on a BEV cell whose warped foreground density exceeds the communication threshold:
\begin{equation}
\displaystyle
\max_{(u,v)\in\bm Q_{i, l = 0}} \bm V_{j\!\to\!i}(u,v) > \tau_{com} ,
\end{equation}
where \((u,v)\) are BEV grid indices.  
The test is performed only at the finest scale \(l{=}0\), whose higher resolution captures the most detailed occupancy information.

\paragraph{Halo-enriched Sparse Feature Encoding.}\ 
Most existing methods~\cite{xu_opv2v_2022,hu_where2comm_2022,qiao2023adaptive,yang_3dssd_2020}
perform early-stage projection: they first transform every CAV’s point cloud into the ego frame and then learn all subsequent features there.
In real-time V2X, however, a vehicle may connect to several neighbours; repeatedly transforming and processing identical point clouds in multiple coordinate frames quickly becomes computationally prohibitive.
Some methods~\cite{shi_pvrcnnpp_2022, lu2023robust} sidesteps this by extracting features in each agent’s own frame and warping them into collaborators’ frames.

Following the feature warping strategy adopted in prior works, we introduce halo enrichment:
Each collaborator first warps its multi-scale feature maps
\(\{\bm F_{j,l}\}\) into the ego frame, then augments every queried grid cell with the features of its \(3{\times}3\) offsets, concatenated along the channel dimension. This enriches the spatial context for subsequent deformable attention, reducing the impact of limited sampling density and providing robustness against the minor pixel-level misalignments that can occur during feature warping.

Finally, only the halo-enriched features at the ego agent’s query locations are transmitted,
\(\{\bm H_{j\!\to\!i,l}\in\mathbb R^{N_l^{q}\times 9C_l}\}\),
providing semantically rich yet bandwidth-efficient inputs for cross-agent fusion.
\subsection{Gated Multi-Scale Deformable Fusion}\label{sec:fusion}

After halo-enriched, sparse features arrive from the selected collaborators,
fusion proceeds in three steps: Step 1: CAV aggregation, Step 2: deformable cross-attention, and Step 3: gated residual blending.

\paragraph{Step 1 — CAV aggregation.}
For each scale \(l\), features from the \(N\) participating CAVs are averaged:
\begin{equation}
\bm{H}_{i,l}^{\text{CAV}} = \frac{1}{N} \sum_{j=1}^{N} \bm{H}_{j \to i, l}
\end{equation}
We use simple averaging as a robust, parameter-free baseline to create a consensus representation, which prevents any single noisy collaborator from dominating the fused feature.

\paragraph{Step 2 — Deformable cross-attention.}
The ego’s query embeddings \(\bm E_{i,l}\), query locations \(\bm Q_{i,l}\),
and the aggregated tensor \(\bm H^{\text{CAV}}_{i,l}\) are passed to a
multi-head deformable-attention module~\cite{zhu2020deformable}.
Each query samples its reference point and learned offsets, producing fused query features that are scattered back to the BEV grid:
\begin{equation}
\bm{F}_{i,l}^{\text{CAV}} = \text{Scatt}\left( \text{DeAttn}\left( \bm{E}_{i,l}, \bm{Q}_{i,l}, \bm{H}_{i,l}^{\text{agg}} \right), \bm{Q}_{i,l} \right)
\end{equation}
with untouched cells filled with zeros.

\paragraph{Step 3 — Gated residual blending.}
Because \(\bm F^{\text{CAV}}_{i,l}\) and the ego features \(\bm F_{i,l}\)
come from different distributions, a spatial gate modulates their mixture:
\begin{equation}
\bm{\tilde{F}}_{i,l} = (1 - \bm{G}_{i,l}) \odot \bm{F}_{i,l} + \bm{G}_{i,l} \odot \bm{F}_{i,l}^{\text{CAV}}
\end{equation}

where
\begin{equation}
\bm{G}_{i,l} = \sigma\left( \text{Conv}_{1\times1}\left[ \bm{F}_{i,l} \,\|\, \bm{F}_{i,l}^{\text{CAV}} \right] \right) \in [0,1]^{C_l \times H_l \times W_l}
\end{equation}
Here \(\parallel\) denotes channel concatenation, \(\sigma\) is the sigmoid,
and \(\odot\) is element-wise multiplication. The \(1{\times}1\) convolution learns to emphasise useful collaborative cues. The rational behind using a gating mechanism is that the aggregated CAV features \( \bm{F}_{i,l}^{\text{CAV}} \) and the original ego features \( \bm{F}_{i,l} \) originate from different sources and exhibit distinct distributions. Directly adding them may disrupt feature consistency. The gating network learns to adaptively balance these two streams, suppressing noise and highlighting useful fused content.

\paragraph{Multi-scale aggregation and detection.}
The collection \( \{ \bm{\tilde{F}}_{i,l} \} \) is forwarded to the same ResNet
aggregation block used in the encoder (\cref{sec:encoder}),
producing a unified BEV tensor \(\bm{\tilde S}_{i}\).
A PointPillars detection head then processes \(\bm{\tilde S}_{i}\) to generate the final 3-D predictions.


\section{Experiment}
\subsection{Experimental Setup}
\label{sec:imp_details}
Our experiments are conducted on the Adver-City-R and OPV2V-R datasets, as introduced in \cref{sec:dataset}. The voxelization settings include a grid size of (0.4\,m, 0.4\,m, 4\,m), a maximum of 32 points per voxel, and a communication range of 70\,m. The detection area is defined as a cuboid with dimensions 281.6\,m (length), 80\,m (width), and 4\,m (height), with each vehicle located at the center of its own detection region.

For our model, we adopt the Adam optimizer with a learning rate of 0.002, a weight decay of \(1 \times 10^{-4}\), and an epsilon value of \(1 \times 10^{-10}\). Gradient clipping is applied with a maximum norm of 1. A MultiStep learning rate scheduler is used with a decay factor \(\gamma = 0.1\), and learning rate drops scheduled at epochs 10 and 15.

We use three scales for cross-agent fusion. The corresponding feature map shapes are \( (128, 100, 352) \), \( (256, 50, 176) \), and \( (512, 25, 88) \), respectively. The number of reference points per scale is set to 200, 100, and 50 for each query generator branch. Each reference point is associated with 9 learnable offsets and 4 attention heads. Offsets that fall on the same location are only transmitted once. Occluder centroids are selected with percentile thresholds $p_l\!\in\!\{0.5, 0.5, 0.5\}$ . 

We adopt the PointPillars detection loss—focal loss for classification (\(\alpha = 0.25,\ \gamma = 2.0\)) and smooth-\(L_{1}\) regression on the 7-D bounding box with \(\lambda_{\text{box}} = 2.0\)—and augment it with an offset-regularization term (see \cref{sec:query_generator}).  
The auxiliary loss is weighted at \(0.1 \times\) the current detection loss and uses an adaptive margin \(\delta_{l}\) proportional to the average occluder-to-shadow distance at each scale.  
This encourages the exploratory branch to learn larger offsets than the heuristic branch, ensuring the two branches maintain distinct behaviours.

All baseline methods are trained using their original configurations, including batch sizes, optimizers, projection strategy, and learning rates. Since no existing method supports LiDAR-Radar fusion for cooperative perception, we implement the same encoder as our model (introduced in \cref{sec:encoder}) for them to ensure a fair comparison. Specifically, a radar voxel feature encoder (VFE) branch is added, and the resulting radar pillar features are concatenated with the LiDAR features before being passed to their corresponding backbones. All models are trained on two RTX 3090 GPUs.

\subsection{Quantitative Evaluation}
To evaluate our proposed framework, we benchmark its performance against several leading collaborative perception methods. \cref{tab:main_results} shows the detection performance across two distinct data splits designed to test the model under varying levels of environmental complexity. The Dense scenarios in AdverCity feature a 2.67x increase in the number of vehicles compared to the Sparse scenarios, thereby significantly amplifying the degree of occlusion~\cite{karvat_adver-city_2025}.
\subsubsection{Bandwidth and Precision Measurement}
During evaluation, each collaborator transmits only its non-zero feature values; zeros are skipped via sparse encoding. As stated in our experimental setup, these evaluations are performed with neither feature compression nor bandwidth limits applied to ensure a fair and direct comparison of algorithmic efficiency.

Therefore, following prior works (e.g., Where2Comm~\cite{hu_where2comm_2022}, BM2CP~\cite{zhao2023bm2cp}, Scope~\cite{yang2023spatio}), we report communication cost as the average per-frame payload assuming float32 features, i.e., 4 bytes per element. The total payload for a given scene $s$ is $B_{s}=4\sum_{l}N_{s,l}$~[bytes], and the average bandwidth reported in our results is $\overline{MB}$~[MB/frame]. This metric represents the raw, uncompressed data payload required by the algorithm before any quantization or channel-specific encoding would be applied in a real-world deployment.
\begin{table}[htbp]
\centering
\scriptsize
\begin{tabular}{ccccc}
\toprule
\multicolumn{5}{c}{\textbf{Adver-City-R}} \\
\midrule
\multirow{2}{*}{Method} & AP@0.5\,$\uparrow$ & AP@0.7\,$\uparrow$ & \multirow{2}{*}{CV\,$\downarrow$} & \multirow{2}{*}{BD\,$\downarrow$} \\
& G. / S. / D. & G. / S. / D. & & \\
\midrule
AttFusion~\cite{xu_opv2v_2022}  & 0.64 / 0.69 / 0.63 & 0.47 / 0.54 / 0.46 & 19.28 & 13.47 \\ 
S-AdaFusion~\cite{qiao2023adaptive}   & 0.66 / 0.71 / \textbf{0.65} & \textbf{0.54} / 0.59 / \textbf{0.52} & 19.52 & 16.14 \\ 
SCOPE~\cite{yang2023spatio}           & 0.38 / 0.43 / 0.36 & 0.26 / 0.33 / 0.24 & 18.9 & 10.14  \\ 
Where2Comm~\cite{hu_where2comm_2022}  & 0.47 / 0.47 / 0.47 & 0.23 / 0.33 / 0.29 & 18.30 & 6.20  \\ 
SlimComm (Ours)                     & \textbf{0.67} / \textbf{0.72} / \textbf{0.65} & \textbf{0.54} / \textbf{0.63} / \textbf{0.52} & \textbf{14.97} & \textbf{1.13}  \\ 
\midrule
\multicolumn{5}{c}{\textbf{OPV2V-R}} \\
\midrule
Method & AP@0.5\,$\uparrow$ & AP@0.7\,$\uparrow$ & CV\,$\downarrow$ & BD\,$\downarrow$ \\
\midrule
AttFusion~\cite{xu_opv2v_2022}  & 0.89 & 0.80 & 19.29 & 6.72 \\
S-AdaFusion~\cite{qiao2023adaptive}   & \textbf{0.91} &\textbf{ 0.85} & 20.47 & 16.35 \\
SCOPE~\cite{yang2023spatio}           & 0.87 & 0.78 & 18.88 & 5.03 \\
Where2Comm~\cite{hu_where2comm_2022}  & 0.86 & 0.77 & 18.74 & 4.45 \\
SlimComm (Ours)                      & 0.87 & 0.80 & \textbf{16.07} & \textbf{0.63} \\
\bottomrule
\end{tabular}
\caption{Detection performance (AP@0.5 / AP@0.7), communication volume (CV, measured in $\log_2$ scale), and bandwidth usage (BD, in MB/frame) across methods at General scenarios. AdverCity results are split into General (G.), Sparse (S.), and Dense (D.) scenarios.}
\label{tab:main_results}
\end{table}
As shown in \cref{tab:main_results}, SlimComm consistently achieves a superior balance between detection performance and communication efficiency across both benchmarks. On Adver-City-R, which is characterized by more complex scenarios with a higher average number of neighboring vehicles per scene, SlimComm delivers state-of-the-art precision. The increased vehicle density and resulting occlusions cause the query-generation mechanism to adaptively increase its bandwidth usage to 1.13 MB to gather the necessary information. This is particularly effective in dense scenarios, where methods like Where2Comm and Scope, lacking semantic prior guidance, fail to capture critical occluded regions. In contrast, SlimComm's occlusion-aware ERP mechanism successfully identifies these areas, matching the accuracy of data-intensive methods while remaining highly efficient.

This efficiency is further highlighted on the OPV2V-R dataset, which has fewer vehicles on average in its test scenarios. Here, SlimComm’s performance remains highly competitive with top-tier methods while requiring only 0.63 MB of bandwidth, an approximate 7x reduction compared to the next most efficient method (Where2Comm at 4.45 MB). This demonstrates that SlimComm’s query-based strategy successfully adapts to scene complexity, preserving high-quality perception while drastically cutting communication overhead and proving its viability for real-world, bandwidth-constrained applications.

\subsubsection{Robustness under Localization and Heading Error }
To assess resilience against localization and heading inaccuracies, we simulate noise on both OPV2V-R and Adver-City-R. Localization noise is injected by adding zero-mean Gaussian perturbations to the ego vehicle’s position, with standard deviations from 0.0\,m to 0.6\,m. Heading noise is modeled by perturbing the yaw angle with standard deviations between 0.0° and 1.0°.

\begin{figure}[h!]
    \centering
    \includegraphics[width=1\linewidth]{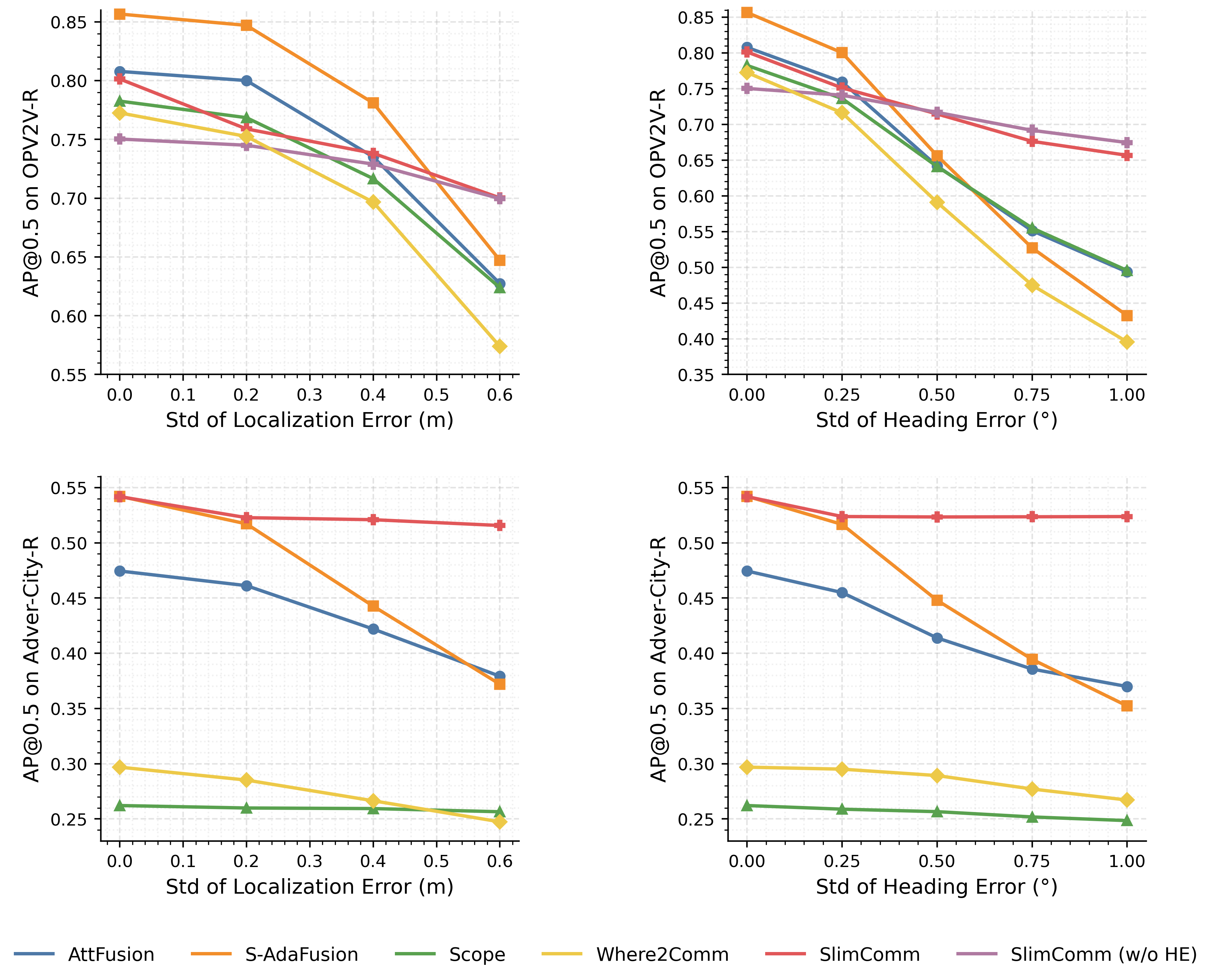}
    \caption{Robustness to localization and heading inaccuracies on top OPV2V-R and bottom Adver-City-R under varying noise levels. Left in each panel: effect of localization noise; right: effect of heading noise.}
    \label{fig:loc_head_robustness}
\end{figure}

As shown in Fig.~\ref{fig:loc_head_robustness} top and bottom, accuracy declines as noise increases, but the rate of degradation varies substantially across methods. Dense-map sharing baselines (AttFusion, S-AdaFusion) start strong at low noise but drop sharply at higher heading errors. Confidence-map-based sparse methods (Where2Comm, Scope) have lower baseline accuracy and are more affected by both noise types. SlimComm maintains the highest AP@0.7 across most noise levels, confirming its robustness to pose perturbations. Interestingly, SlimComm (w/o HE) matches or slightly surpasses SlimComm in high-noise regimes, suggesting that halo enrichment, while beneficial for precision in nominal settings, can be less advantageous under severe localization or heading errors, as the larger receptive region amplifies the influence of misaligned information.

We further evaluate robustness under asynchronous message delays, where feature exchange is shifted by 200--600~ms. SlimComm maintains stable performance with less than 1.5 mAP degradation, whereas baseline methods degrade more significantly. 
Detailed delay-performance curves are provided in the supplementary material.

\subsection{Ablation Studies}\label{sec:ablation}
\paragraph{Impact of Different Components.}
We conduct ablation studies to evaluate the contribution of each component in our framework, as shown in ~\cref{tab:module_ablation}. On Adver-City-R, introducing the exploratory branch improves the AP@0.5 from 0.58 to 0.61 and AP@0.7 from 0.41 to 0.49, demonstrating the benefit of explicitly exploring occluded or uncertain areas. Further incorporating the halo-enrichment mechanism brings an additional performance gain, reaching AP@0.5 of 0.67 and AP@0.7 of 0.54, which confirms that using spatial context is an effective strategy to mitigate feature warping distortion and sample sparsity. On OPV2V-R, a similar trend is observed.
\begin{table}[ht]
\centering
\scriptsize
\begin{tabular}{ccccccc}
\toprule
\multicolumn{3}{c}{\textbf{Module}} & \multicolumn{2}{c}{\textbf{Adver-City-R}} & \multicolumn{2}{c}{\textbf{OPV2V-R}} \\
\midrule
{HRP} & {ERP} &{HE} &  AP@0.5\,$\uparrow$ & AP@0.7\,$\uparrow$ & {AP@0.5\,$\uparrow$} &{AP@0.7\,$\uparrow$} \\
\midrule
\checkmark &            &            & 0.58  & 0.41 & 0.83 & 0.73 \\
\checkmark & \checkmark &            & 0.61 & 0.49 & 0.86 & 0.75 \\
\checkmark & \checkmark & \checkmark & 0.67 & 0.54  & 0.87 & 0.80\\
\bottomrule
\end{tabular}
\caption{Ablation study on different module combinations. HRP: Heuristic Branch in Query Generator, ERP: Exploratory Branch in Query Generator, HE: Halo-enrichment}
\label{tab:module_ablation}
\end{table}
\paragraph{Impact of Reference Points.}
An ablation study was conducted to investigate how the number of queries affects performance and communication cost, as shown in \cref{tab:query_ablation}. These queries are initialized from a set of reference points derived from semantic priors. Using a minimal set of (50, 25, 15) reference points per scale leads to the lowest detection accuracy, indicating insufficient coverage of critical object and occlusion regions. Increasing the number to (100, 50, 25) and further to the default setting of (200, 100, 50) progressively improves both AP@0.5 and AP@0.7. This confirms the trade-off between accuracy and bandwidth: while more queries enhance perception quality, they also increase communication cost. Considering this balance, the (200, 100, 50) setting was adopted for our main experiments.
\begin{table}[h!]
\centering
\scriptsize
\begin{tabular}{ccccc}
\toprule
\multicolumn{5}{c}{\textbf{Impact of RP Numbers on Adver-City-R}}   \\
\midrule
RP &  AP@0.5\,$\uparrow$ & AP@0.7\,$\uparrow$ & {CV\,$\downarrow$} & BD\,$\downarrow$ \\
\midrule
(50, 25, 15)  & 0.59  & 0.48 & 12.81 & 0.29 \\
(100, 50, 25) & 0.61 & 0.48 & 14.22 & 0.65 \\
(200, 100, 50)* & 0.67 & 0.54  & 14.97 & 1.13 \\
\midrule
\multicolumn{5}{c}{\textbf{Impact of Communication Threshold $\tau_{com}$ on OPV2V-R}} \\
\midrule
$\tau_{com}$ & AP@0.5\,$\uparrow$ & AP@0.7\,$\uparrow$ & CV\,$\downarrow$ & BD\,$\downarrow$ \\
\midrule
0*  & 0.87 & 0.80 & 16.07 & 0.63 \\
0.25   & 0.85 & 0.76 & 15.53 & 0.61 \\
0.50          & 0.84 & 0.75 & 13.15 & 0.52 \\
0.75  & 0.83 & 0.73 & 9.72 & 0.38 \\
\bottomrule
\end{tabular}
\caption{Analysis of the impact of RP numbers and communication threshold $\tau_{com}$. Default settings are marked with an asterisk (*).}
\label{tab:query_ablation}
\end{table}

\paragraph{Impact of Communication Threshold.}
To evaluate the impact of our collaborator-selection mechanism, we sweep the communication threshold~$\tau_{com}$ on the OPV2V-R validation split (see \cref{tab:query_ablation}). As $\tau_{com}$ increases from 0 to 0.75, communication volume (CV) and bandwidth (BD) decline steadily, reflecting the pruning of low-value collaborators, while detection accuracy also drops. We select $\tau_{com}=0$ in our main experiments, which offers a favorable balance by removing only the most uninformative links while preserving peak perception performance.

\section{Limitations and Future Work}\label{sec:future}

Although SlimComm advances bandwidth-aware cooperative perception, several avenues remain open. A critical next step is real-world validation. All current experiments rely on CARLA because no public multi-agent radar dataset with per-point Doppler is yet available. SlimComm will be re-trained and evaluated as soon as such data are released (e.g., V2X-Radar~\cite{yang2024v2x}).

A key deployment challenge is handling communication delay, packet asynchrony, and imperfect localization, which induce feature-warping misalignment. These effects are not explicitly modeled in the current design. Robustness to localization and heading errors is reported; nevertheless, online pose-refinement and automatic correction within the fusion module remain to be integrated. Future work will introduce synchronization mechanisms and feature-alignment techniques such as spatial cross-attention or offset correction, and the intrinsic robustness of halo enrichment to pixel-level warps will be examined.

To narrow the sim-to-real gap in the interim, higher-fidelity radar simulation is planned. The present simulator approximates multipath and ghost targets; a more advanced engine (e.g., C-Shenron~\cite{srivastava_realistic_nodate}) will be incorporated, and an updated \emph{OPV2V-R$^{+}$} split with richer sensor physics will be released.

Further research will strengthen LiDAR–radar fusion beyond simple concatenation, learn adaptive query budgets conditioned on scene complexity and link congestion, and provide Doppler-free fallbacks using optical or LiDAR scene-flow surrogates. Occlusion-aware queries also lend themselves to motion forecasting; extending SlimComm to joint detection–prediction is natural. Although sparser than full maps, queries may still leak location cues and should be assessed accordingly.

\section{Conclusion}

We presented SlimComm, a proactive, query-driven framework that unites LiDAR and 4-D radar for bandwidth-efficient cooperative perception.  
Guided by motion-centric and confidence-based priors, SlimComm drops sparse queries on dynamic objects and occlusion shadows, prompting neighbours to return only the most informative BEV features.  A gated multi-scale deformable-attention block then fuses ego and halo-enriched neighbour features. Experiments on the new OPV2V-R and Adver-City-R benchmarks show that SlimComm matches full-map sharing while transmitting just $\sim$10\,\% of the data, and it consistently outperforms prior sparse-communication baselines.  
Remaining challenges, such as richer LiDAR–radar coupling, delay-aware fusion, and real-world evaluation, are outlined in Section~\ref{sec:future}. Addressing them will bring SlimComm closer to deployment in next-generation V2X perception.

\paragraph{Acknowledgements.} This paper was created in the Country 2 City - Bridge project of the German Center for Future Mobility, which is funded by the German Federal Ministry for Digital and Transport.

\clearpage
\twocolumn[
\begin{center}
    {\Large \bf Supplementary Material \par}
    \vspace{1ex}
    {\normalsize SlimComm: Doppler-Guided Sparse Queries for Bandwidth-Efficient Cooperative 3-D Perception \par}
    \vspace{2ex}
\end{center}
]

\section{Dataset Details}

\subsection*{Vehicles in Ego Perception Range}

\begin{figure}[h!]
  \centering
  \includegraphics[width=1\linewidth]{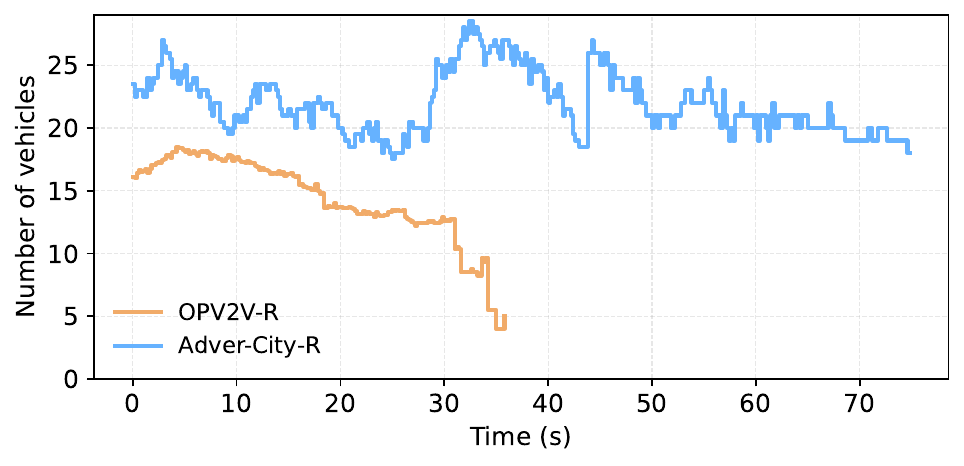}
  \caption{Number of vehicles within the perception range of the ego vehicle in the OPV2V-R and Adver-City-R test sets.}
  \label{fig:PerceptionRange}
\end{figure}

Figure~\ref{fig:PerceptionRange} shows the number of vehicles within the perception range of the ego vehicle. 
On average, OPV2V-R contains 16.74 $\pm$ 7.90 vehicles per frame, while Adver-City-R 
has 20.75 $\pm$ 7.28. This highlights that Adver-City-R exhibits higher traffic density, 
which increases the likelihood of occlusions and makes perception more challenging.  
\subsection*{Scenario Statistics in Adver-City-R}

As Table~\ref{tab:adver_city_stats} shows, Adver-City-R covers five distinct 
scenario types that reflect common driving situations in both rural and 
urban environments. Scenario lengths range from about 23\,s to 39\,s, 
providing clips of manageable duration while still capturing diverse 
traffic interactions. The traffic density differs strongly across road types: 
rural scenarios typically involve around 10 vehicles within the ego 
perception range, whereas urban intersections exceed 30 on average. 
Speed distributions also vary: vehicles and CAVs travel faster in rural 
settings, while urban intersections are characterized by slower and 
denser traffic flows. These variations emphasize the heterogeneity of 
Adver-City-R and demonstrate its suitability for evaluating perception 
systems under diverse and challenging traffic conditions.  

\subsection*{Road Type Splits in Adver-City-R and OPV2V-R}

Adver-City-R enforces a strict road-type split: 
\textit{urban intersections} occur only in the test set, 
\textit{rural curved non-junction} roads only in validation, 
and \textit{rural intersections}, \textit{rural straight non-junction} roads, 
and \textit{urban non-junction} roads only in training. 
This means that test scenarios involve road geometries that are unseen during training, 
leading to a more demanding evaluation. Combined with the generally higher traffic 
density and occlusion levels of Adver-City-R, this results in more challenging 
conditions compared to OPV2V-R. In contrast, OPV2V-R does not enforce such separation, 
and all road types appear across its splits.

\begin{figure}[t]
  \centering
  \begin{subfigure}[t]{\linewidth}
    \centering
    \includegraphics[width=\linewidth]{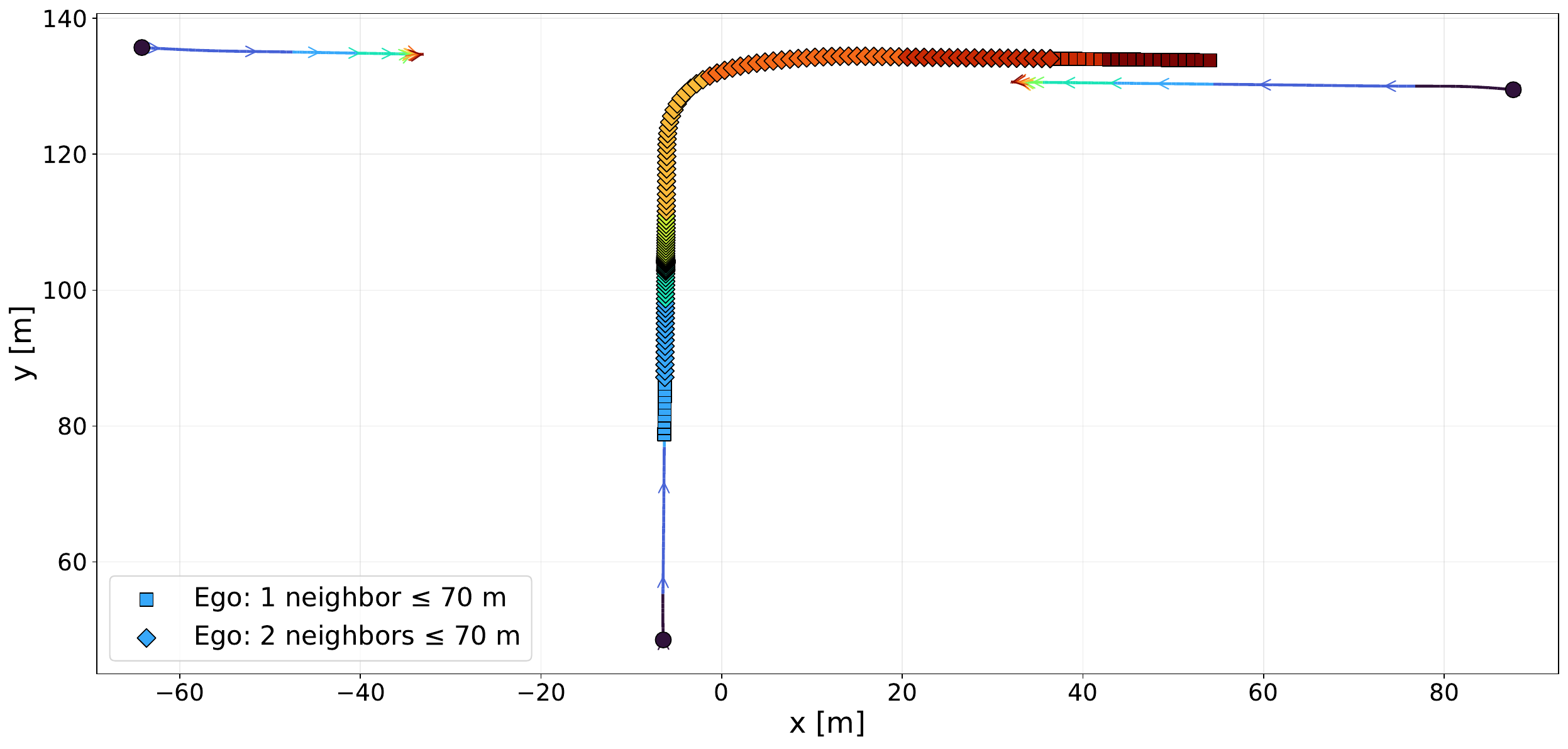}
    \caption{Adver-City-R}
  \end{subfigure}

  \vspace{0.4em}

  \begin{subfigure}[t]{\linewidth}
    \centering
    \includegraphics[width=\linewidth]{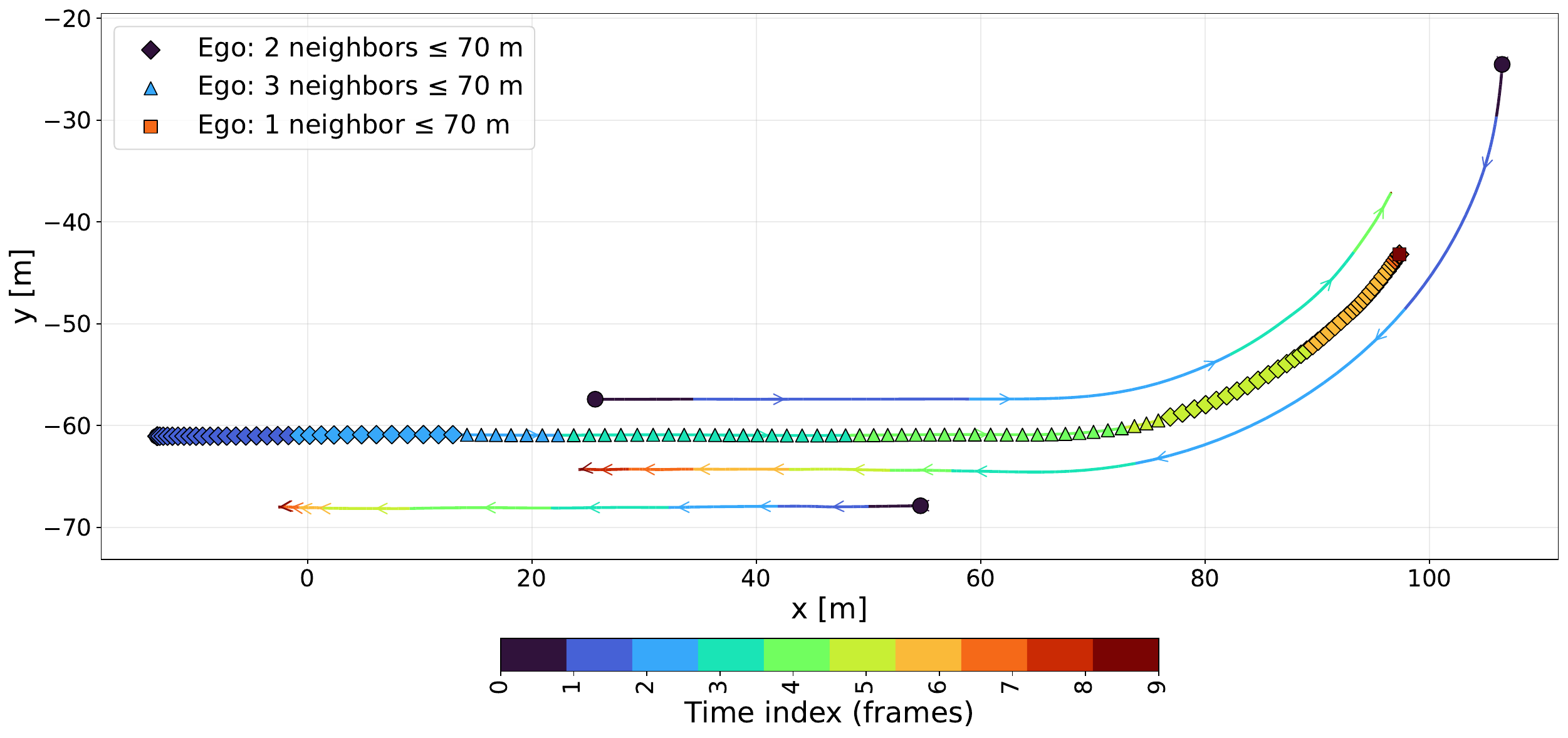}
    \caption{OPV2V-R}
  \end{subfigure}

  \caption{Ego vehicle routes and communication events in OPV2V-R and Adver-City-R. 
Trajectories are color-coded by time; markers indicate the number of neighbors within 70\,m.}

  \label{fig:routes_comm}
\end{figure}

\begin{table*}[h!]
\centering
\scriptsize
\begin{tabular}{lccccc}
\toprule
Scenario & Percentage(\%) &
\begin{tabular}[c]{@{}c@{}}Length(s)\\ mean / std\end{tabular} &
\begin{tabular}[c]{@{}c@{}}Traffic density\\ mean / std\end{tabular} &
\begin{tabular}[c]{@{}c@{}}Traffic Speed (km/h)\\ mean / std\end{tabular} &
\begin{tabular}[c]{@{}c@{}}CAV speed (km/h)\\ mean / std\end{tabular} \\
\midrule
rural\_curved\_non\_junction    & 17.36 & 24.50 / 3.30 & 15.00 / 6.00 & 14.48 / 11.45 & 28.45 / 10.29 \\
rural\_intersection            & 27.28 & 22.80 / 0.00 & 10.00 / 4.00 & 15.64 / 11.37 & 10.36 / 12.95 \\
rural\_straight\_non\_junction  & 16.16 & 38.50 / 2.60 & 10.00 / 4.00 & 21.86 / 11.24 & 21.31 / 13.11 \\
urban\_intersection            & 18.14 & 29.70 / 7.80 & 33.00 / 15.00 & 12.92 / 12.72 & 10.95 / 11.32 \\
urban\_non\_junction           & 21.05 & 25.60 / 1.10 & 27.00 / 12.00 & 13.69 / 11.96 & 25.58 / 12.67 \\
\midrule
\textbf{overall}               & 100.00 & 28.20 / 6.90 & 19.00 / 13.00 & 15.28 / 12.37 & 19.37 / 14.10 \\
\bottomrule
\end{tabular}
\caption{Summary of Adver-City-R dataset statistics by road type. 
Traffic density means the number of vehicles spawned around the ego
vehicle within a 140\,m radius}
\label{tab:adver_city_stats}
\end{table*}
Figure~\ref{fig:routes_comm} shows the trajectories of all vehicles 
in representative scenarios from OPV2V-R and Adver-City-R. The routes are color-coded with a time gradient from the beginning to the end of each scenario. 
Markers highlight the frames in which the ego vehicle detects neighboring vehicles within the 
\textbf{70\,m communication range}. The marker shape indicates the number of neighbors: 
a square represents one, a diamond two, a triangle three, and a star four or more. 

A clear structural difference can be observed between the two datasets. In OPV2V-R, 
vehicles frequently move along the same road segments, resulting in long periods of 
continuous communication. In contrast, Adver-City-R features vehicles approaching from 
different directions and meeting mainly at intersections, which concentrates communication 
events around junction areas.



\section{Ego Query Generator Architecture}
\begin{figure}[h]
    \centering
    \includegraphics[width=1\linewidth]{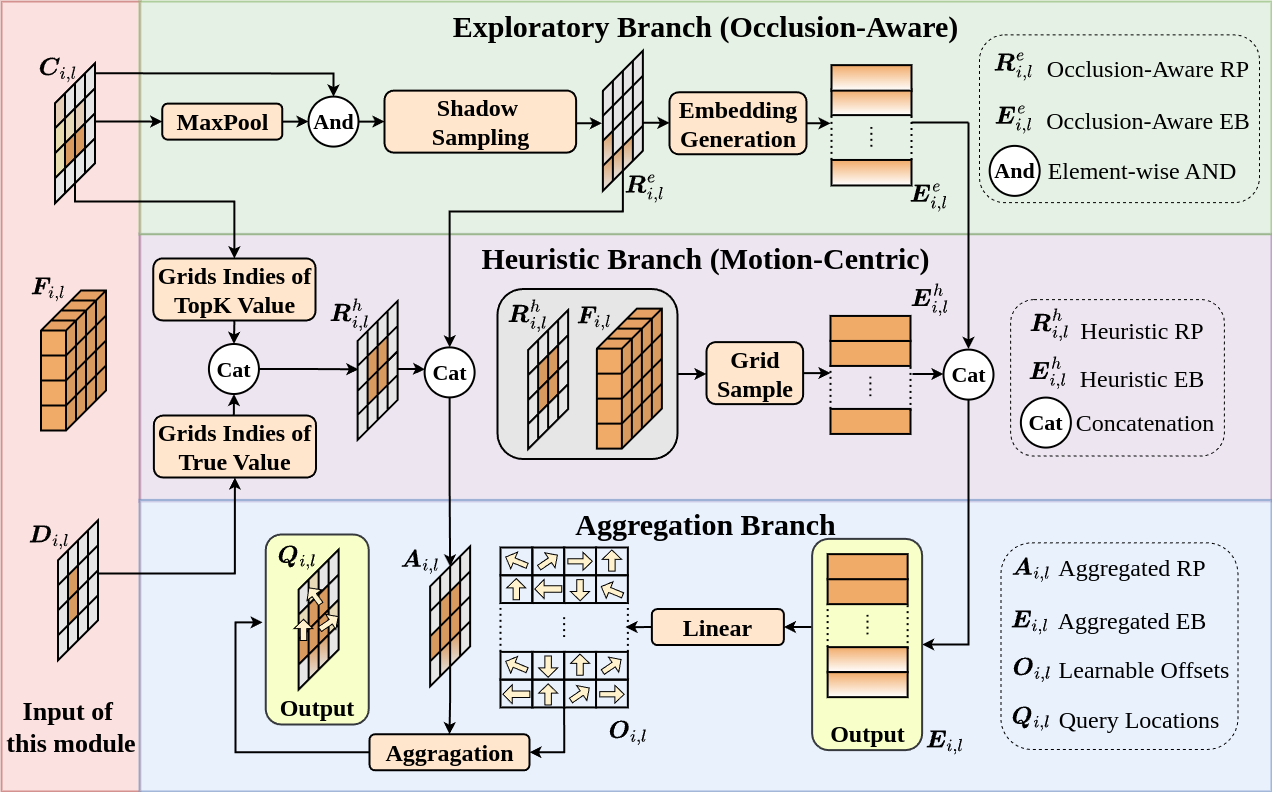}
    \caption{Detailed architecture of the Ego Query Generator, consisting of a motion-centric heuristic branch, an occlusion-aware exploratory branch, and an aggregation branch that combines reference points (RP) and embeddings (EB) into the final query set.}
    \label{fig:query}
\end{figure}
Figure~\ref{fig:query} gives a compact overview of Ego Query Generator.
Per scale, the module (i) selects HRP from dynamic cells and top-confidence cells,
(ii) places ERP behind occluder peaks in the confidence map via shadow sampling, and
(iii) concatenates both and applies a coarse offset followed by a $3\times3$ deformable halo to obtain the final sampling locations.

\section{Qualitative Evaluation}
\subsection{Quality of RP and Query Locations}
\begin{figure}[h!]
  \centering
  \includegraphics[width=1\linewidth]{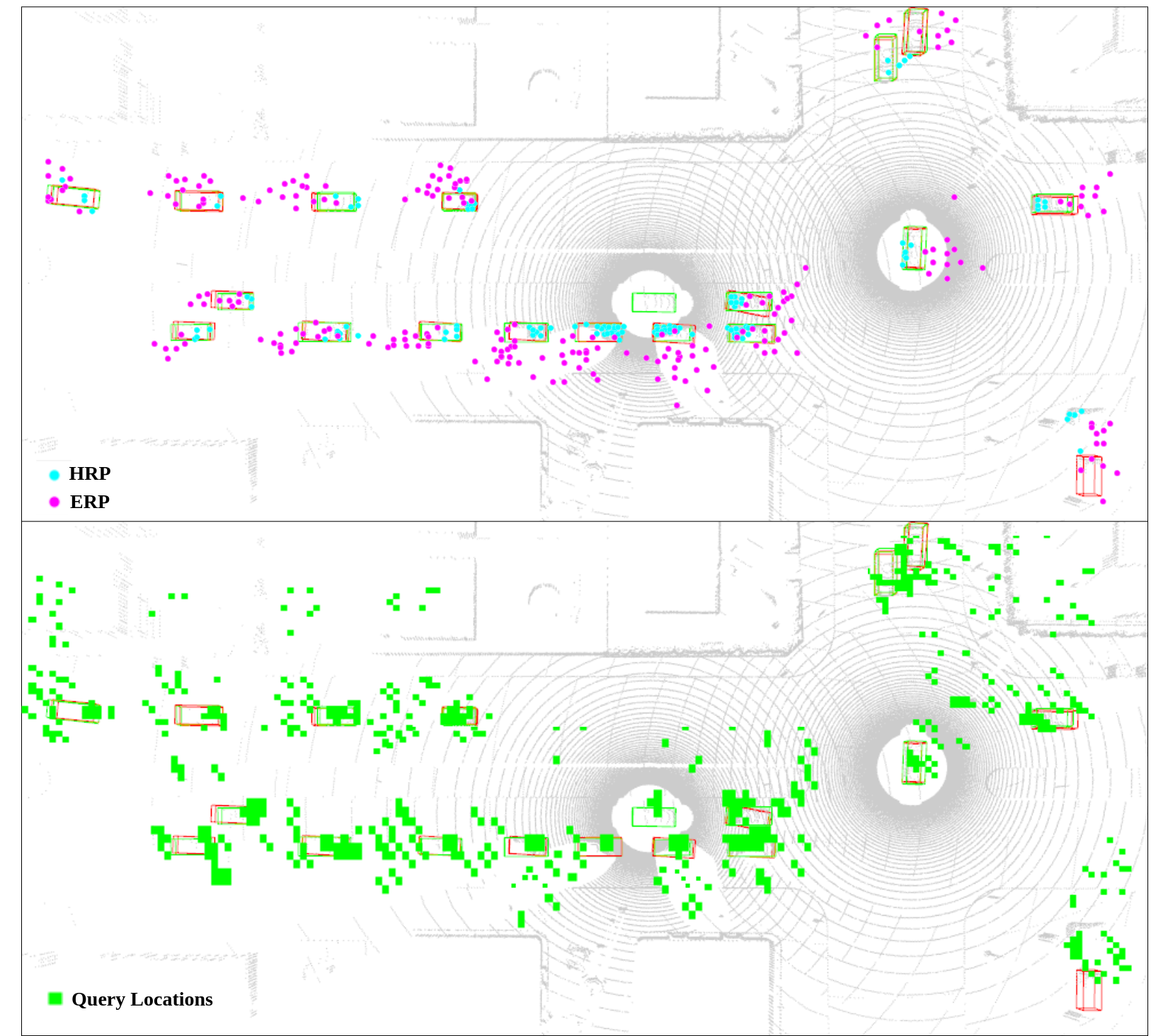}
    \caption{Qualitative distribution of HRP (cyan), ERP (magenta), and final query locations (green) with $3\times3$ halos in a dense scene.}
  \label{fig:hrp_erp}
\end{figure}
Figure~\ref{fig:hrp_erp} shows complementary spatial patterns of the two reference points types and final query locations.  As shown in the top panel, HRP concentrate on visible vehicles and high-confidence edges, reflecting the motion/score-driven selection used for object refinement. ERP are cast behind strong returns along the ego-to-occluder rays, spreading into partially or fully occluded regions.
This complementarity expands coverage into occlusions while keeping precision on visible objects, which translates to fewer misses in cluttered areas without introducing extra false positives. 

The bottom panel visualizes the final query locations. They are obtained by
applying the two-stage offset to the union of HRP/ERP anchors: a coarse nudge to the
anchor center followed by a $3{\times}3$ deformable halo. As a result, queries appear as
compact $3{\times}3$ clusters around visible vehicles and extend into the occluder
shadow regions suggested by ERP. Background road areas remain sparsely sampled. This
distribution concentrates sampling where evidence is strongest or likely hidden, while
keeping the overall query budget low. 
\subsection{Qualitative Comparison with SOTA Methods}
\begin{figure}[h]
    \centering
    \includegraphics[width=1\linewidth]{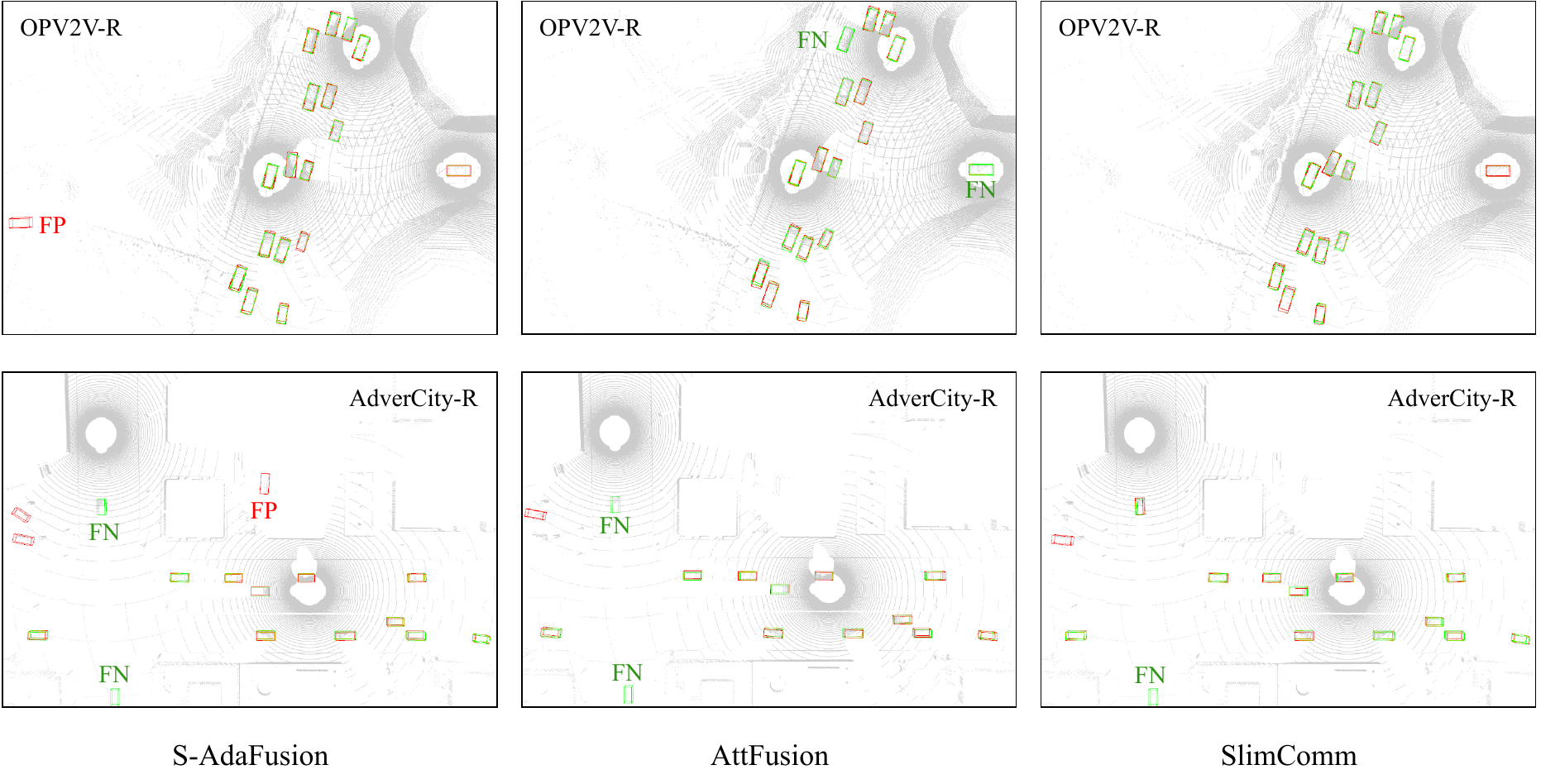}
    \caption{Visualization of detection results in two very dense scenarios. Green: ground truth bounding box. Red: prediction result.}
    \label{fig:qualitative}
\end{figure}
\cref{fig:qualitative} compares S-AdaFusion, AttFusion, and SlimComm on two occlusion-heavy scenes (OPV2V-R and AdverCity-R). As annotated in the figure, in OPV2V-R, S-AdaFusion covers most vehicles but produces a clear FP in clutter. AttFusion misses multiple vehicles near occluders (several FNs). SlimComm successfully detects all vehicles with precise and well-aligned bounding boxes. In AdverCity-R, under stronger occlusion, S-AdaFusion shows both FP and FN, and AttFusion accumulates additional FNs. SlimComm maintains the best alignment and the fewest total errors among the three, reflecting a stronger precision–recall balance under heavy occlusion.

\section{Robustness against Asynchronous}
\begin{figure}[h!]
    \centering
    \includegraphics[width=\linewidth]{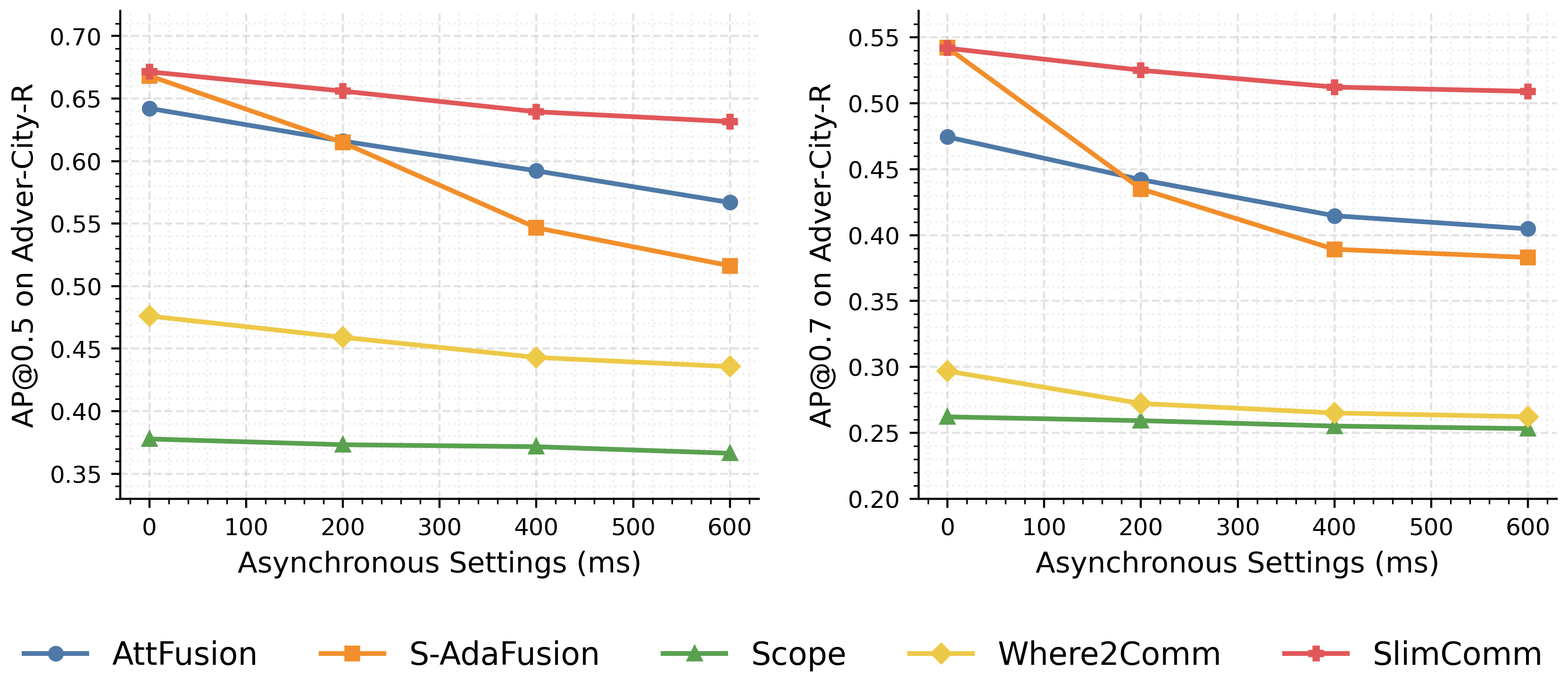}
    \caption{Robustness of different frameworks under localization asynchrony. Left: AP evaluated at IoU = 0.5. Right: AP evaluated at IoU = 0.7.}
    \label{fig:async_robustness}
\end{figure}
Figure~\ref{fig:async_robustness} evaluates methods under injected delays up to 600\,ms on Adver-City-R.
Two patterns emerge: (i) models that maximize dense cross-attention at 0\,ms achieve higher initial AP but are brittle to delay (steep negative slope); (ii) confidence/selection-driven schemes in Where2Comm and Scope transmit only high-confidence regions and therefore exhibit slower degradation.
SlimComm inherits the benefits of both: reference/exploratory queries are placed from stable priors (Doppler motion and foreground confidence), while offset-based deformable attention and halo aggregation reduce reliance on exact temporal alignment.
Consequently, SlimComm maintains a substantially flatter AP–delay curve compared to attention-heavy baselines while retaining high absolute AP.

{
    \small
    \bibliographystyle{ieeenat_fullname}
    \bibliography{main}
}

\end{document}